\documentclass[smallextended]{svjour3}       
\smartqed  
\usepackage{graphicx}
\usepackage[numbers,sort&compress]{natbib}
\usepackage[colorlinks=true,linkcolor=blue,urlcolor=blue,citecolor=blue,anchorcolor=blue]{hyperref}
\usepackage{url}
\begin{document}

\title{Genetic Programming Theory and Practice: A Fifteen-Year Trajectory\footnote{This work was supported by National Institutes of Health grants AI116794, DK112217, ES013508, HL134015, LM010098, LM011360, LM012601, and TR001263.
This is a post-peer-review, pre-copyedit version of an article published in \textit{Genetic Programming and Evolvable Machines}. The final authenticated version is available on the journal’s
website: \url{https://link.springer.com/article/10.1007/s10710-019-09353-5}.}
}

\titlerunning{Genetic Programming Theory and Practice: A Fifteen-Year Trajectory}        

\author{Moshe Sipper         \and
        Jason H. Moore 
}


\institute{Moshe Sipper \at
              Institute for Biomedical Informatics (IBI), Perelman School of Medicine,
University of Pennsylvania, Philadelphia, PA 19104 and Department of Computer Science, Ben-Gurion University, Beer-Sheva 8410501, Israel. \\
              \email{sipper@gmail.com}           
           \and
           Jason H. Moore \at
              Institute for Biomedical Informatics (IBI), Perelman School of Medicine,
University of Pennsylvania, Philadelphia, PA 19104
}

\date{Genetic Programming and Evolvable Machines (2020) 21:169–179}

\maketitle

\begin{abstract}
The GPTP workshop series, which began in 2003, has served over the years as a focal meeting for genetic programming (GP) researchers. As such, we think it provides an excellent source for studying the development of GP over the past fifteen years. We thus present herein a trajectory of the thematic developments in the field of GP.
\end{abstract}

\section{Introduction}
\label{intro}
The \textit{Genetic Programming Theory and Practice (GPTP)} workshop series was launched in 2003 as an annual event intended to promote the exchange of research results between those who focus on genetic programming (GP) theory and those who focus on GP applications. Throughout the years it has maintained its format---a small number of participants and speakers (perhaps two dozen) who engage in deep, lively discussions---as well as its main focus on GP. Participation is mostly by invitation only, 
with the annual organizers attempting to identify key players, who are then treated to a less stringent review process, allowing them to push the boundaries and speculate further than in other venues. The annual proceedings present a wealth of research, and many of them contain a summary chapter with a rare treasure trove in science: an account of oral discussions. 
All these characteristics set GPTP apart from other conferences in the field, such as EuroGP and GECCO.

In the preface to the fourteenth event, \citet{Tozier:2016preface} averred the workshop's importance and influence:
``The central and explicit focus of the GPTP Workshop has
always been the conversations that are fostered \textit{at} the meeting itself \ldots As a result, the proceedings volume from each workshop should only ever be read as a record of where some subset of attendees started individually, not as a position in which we ended up as a group by the end. The \textit{real} results will only gradually appear in the subsequent literature, as a cloud of works that may not even explicitly refer to this meeting, and in the subtly-changed directions of ongoing research programs in years to come.''

Given the workshop's focus, scope, and nature, we think it can serve as the perfect medium for an inquiry into the development of GP over the years. Our aim herein is to present the thematic development of the field of GP, over a 15-year period, as seen through the lens of GPTP---from GPTP I to GPTP XV. We do not aim to narrate a detailed overview of research, nor delve into the technical aspects of the 200-plus papers presented in total. Rather, our goal is to discuss thematic trends, analyzing, as it were, the trajectory of GP research.  

\section{From I to XV}

\paragraph{GPTP I: 2003.}
\citet{worzel:2002} wrote in the inaugural volume: ``The dynamics of the GP process is poorly understood with many serious questions remaining. People applying GP to real-world problems have relied more on intuition than theory, experience more than mathematics. To reach the next stage in its development, GP theory and practice must both advance. Theory must inform practice and practice must test theory.''

In the first workshop there seemed to be a rather wide divide between theoreticians and practitioners, the former aiming for simple models that could be analyzed thoroughly, the latter content with viewing success as its own justification \cite{worzel:2002}. 

On the theory side, a main question asked was whether genetic algorithm (GA) theory, e.g., schema theorems, could be extended to GP. Also discussed were questions regarding the role of building blocks, population size, selection operators, and diversity. On the application side, the range was quite wide: analyzing biological data, software engineering, modeling chemical processes, evaluating oilfields, and financial modeling. 

An effervescent discussion of interest was about the use and relevance of biological models, showing the fruitfulness of building bridges between biology and evolutionary computing.
All told, this inaugural meeting was successful in bringing theory and practice to the (literally) same table. 

\paragraph{GPTP II: 2004.} 
Summarizing the second workshop, \citet{oreilly:2004} began with a wonderful quote---still very relevant today---by keynote speaker Dave Davis, who warned that when selling an EC solution it is wrong to state that the algorithm ``is random, and can be run as long as you want to get better solutions.'' Instead, explain that the algorithm ``is creative and can be used to improve on existing solutions.'' Davis also opined that ``the jury is out on the usefulness of theory to applications people'', a sentiment some might still share today.

The workshop consisted again of a mix both of application and theory papers. As with the first workshop, \citet{oreilly:2004} noted the widening gap between theory and practice, the former studying fairly simple, abstract models far less complex than the systems used by the latter. However the interaction of both groups during the workshop led them to optimistically state that, ``despite the current state of theory is imperfect to precisely describe the dynamics of GP and to strictly guide the use of GP \ldots there will always be links between theory and
practice.'' 

\paragraph{GPTP III: 2005.} \citet{yu:2005} noted that the third workshop saw some signs of convergence: papers presenting techniques on small-scale problems discussed how they might be applied to real-world problems, and real-world problems were tackled through insight from theory.

An interesting development in this and the previous year was the appearance of human-competitive results produced by GP (lenses, antennas, analog circuit design). Such results  would grow significantly in number in the years to follow, to be showcased in the HUMIES competition \cite{Kannappan:2015}. Talk of GP as an ``invention machine'' no longer seemed far fetched.
The participants agreed, interestingly, that fields such as antenna design were not yet mature enough to be less accepting of new ideas and were thus more open to GP. 

A prime topic of interest in this workshop was visualization---of the evolutionary process, of results, etc.---which has since garnered yet more attention. Several open challenges were enumerated by the participants, including:
open-ended evolution; the successful use of GP by large teams in industry; handling large datasets; analysis of GP systems; GP integration with other techniques; theoretical tools for understanding large modular systems; using tools from other fields to enhance GP; better visualization infrastructure; more complicated fitness functions; aiming at ``real'' AI goals; integration of domain knowledge; making GP attractive to industry. Though progress has been made along these fronts, many---if not most---of these challenges still apply today.

\paragraph{GPTP IV: 2006.}
\citet{soule:2006} began by restating the workshop's original goal: ``to allow theory to inform practice and practice to test theory.'' Progress seemed to have been made in that ``it was clear that GP's ability to solve real world problems was broadly accepted, but that often this success was not, and could not be, accomplished by
a `pure' GP system.'' This emerging observation persists to this day, as any GP practitioner will attest. Indeed, at the time, \citet{soule:2006} called this move from pure GP ``a significant shift in the field''.
One recurring theme in the workshop was thus the  additions and modifications to vanilla GP needed to solve particular classes of real-world problems. 

Many hurdles that needed to be overcome were observed to be shared by multiple fields of application and thus represented important areas of future research. Interestingly, the theoreticians were seen to mirror the applied papers, studying how various techniques could be combined with GP.

The application areas fit into four areas: robotics, bioinformatics, symbolic regression, and design.These areas are still highly relevant over a decade later. Common hurdles (probably familiar to many readers) were recognized in practice, including: the dimensionality of the search space, the time required for an individual's fitness evaluation, and the need for robust solutions. Workshop participants repeatedly chose a small number of techniques to improve GP: fitness approximation, solution caching, parameter optimization, cooperation (e.g., between individuals), Pareto optimization, and pre- and post-processing. A major success of (non-pure) GP was noted in the field of financial analysis.

A lively discussion about packaging GP for the non-expert concluded that a simple GP engine would be counterproductive, and the favored alternative was to build GP into domain-specific software.

\paragraph{GPTP V: 2007.}
\citet{soule:2007} stated that GP had produced many impressive success stories but that, ``these successes are still generally achieved by researchers with significant expert knowledge of GP; successful application of GP to large-scale problems remains out of the reach of novices.'' The fifth workshop focused on two main issues: understanding the foundations of GP, and the development of advanced techniques for real-world problems. Much discussion emanating from these topics concerned the future of GP.

Regarding the theoretical side, several papers attempted to push the envelope, recognizing that GP was still not well understood. Also addressed was the question of how closely GP should be to what was known about biological evolution. The existence of a developmental phase in nature as opposed to GP is a major difference, and the genotype-phenotype distinction was brought to the fore.

A number of advanced techniques for improving GP (some encountered above) received much attention, including: fitness- and age-layered populations, code reuse through caching and archives, Pareto optimization, pre- and post-processing, and the use of expert knowledge. 
These trends, also explored in the previous workshop, underscored the need for non-vanilla GP when tackling real-world problems. 

Once again concern was expressed regarding the use of GP by non-experts. 
However, a sentiment shared by most attendees was that given the state of the field at the time, building a GP system for a specific application domain, without trying to make it work independently of an experienced user, was the most promising approach for complicated real-world problems. This observation is probably still true today.

\paragraph{GPTP VI: 2008.}
\citet{soule:2008} began by noting GP's progress from a young art toward a mature engineering discipline, evoking the by-now repeating theme of creating an easy-to-use, turnkey GP system for non-experts. Indeed, keynote speaker Marc Schoenauer's spoke of his goal as an evolutionary researcher ``to become extinct'' by supplying GP novices with the tools to apply GP to practical problems without the need for a GP expert to design the system. As noted, we are not there yet.

The participants of this sixth workshop identified a number of critical themes for achieving reliable, industrial-scale GP: making efficient and effective use of test data, reusing subsolutions, increasing the role of a domain expert, and sustaining long-term evolvability (successfully exploring unexplored regions of the search space and avoiding premature convergence).
These themes were observed to be highly inter-related, e.g., ineffective use of test data could slow evolution.

Many practitioners were seen to use a ``kitchen sink'' approach, continually adding many of the aforementioned techniques (Pareto optimization, age layering, etc.). ``This approach is often successful and may be necessary given our still limited knowledge regarding what features make a specific problem difficult and what techniques most effectively address those features. However, it seems unlikely that the kitchen sink approach is the most efficient.'' \cite{soule:2008}

\paragraph{GPTP VII: 2009.}
\citet{Riolo:2009} identified a number of GP challenges, some of which we have encountered before: economic resource usage (processor, memory, disk), ensuring quality results that are usable (e.g., human-interpretable, trustworthy, or predictive on very different inputs), reliable convergence, and addressing the wide range of problem domains (representations, operators, and so forth).

One trend noted in this seventh edition of GPTP was that fewer ``conventional'', pen-and-paper theory submissions were received. The theory was in the form of techniques that were measurably better, easier to analyze, and better explained; these kinds of results promoted a general \textit{best-practice} approach. For example, four papers discussed how fitness-function design was key in making each respective problem tractable for GP. Other papers presented best practices in representation, operator design, and selection.

GPTP VII showed that GP could be used as a discovery engine, e.g., capturing the dynamics in classical physics models such as pendulums. Practitioners showcased time-series modeling, high-dimensional symbolic regression and classification, and financial applications. ``The consensus among the participants this year,'' wrote \citet{Riolo:2009}, ``was that genetic programming has reached a watershed in terms of practicality for a well-defined range of applications.'' This is not quite straightforward because it necessitates an ``appropriate determination of algorithm techniques, representation, operators, and fitness function.'' \cite{Riolo:2009}

The GPTP events were seen to promote successful techniques for alleviating commonly occurring problems such as premature convergence, bloat, and scalability. The focus on best-practice approaches presented a new direction for GP. 

\paragraph{GPTP VIII: 2010.}
\citet{McConaghy:2010} wrote that the workshop was marked by researchers beginning to aim for the next level, that of ``\textit{systems} where GP algorithms play a key role.'' There was a record number of demos presented and an emphasis on system usability and user control. As before, major challenges encountered included efficient use of computational resources, ensuring quality results, and reliable convergence. To these were now added usability goals: ease of system integration, end-user friendliness, and user (interactive) control of the problem.

Once again, GP best practices were pivotal to the workshop, showing how GP could be adapted to solve highly complex problems (e.g., the first chapter describes  best-practice operator design for evolving Java programs \cite{Orlov:2010}).

An interesting observation made by \citet{McConaghy:2010}:
``What is equally significant in these papers is that which is not mentioned or barely mentioned: GP algorithm goals that have already been solved sufficiently for particular problem domains, allowing researchers to focus their work on the more challenging issues.'' 
For example, previously urgent problems such as bloat and interpretability were now dealt with through various means (e.g., Pareto optimization, abstract expressions). This demonstrated clear and promising evidence of progress in the field of GP. Also, an informal survey of the participants showed that most used compute clusters, and that the cloud was starting to gain ground.

Perhaps the most telling statement made this year was: ``The toy problems are gone; the GP systems have arrived.'' On the downside, \citet{McConaghy:2010} noted that GP did not demonstrate the potential associated with natural evolution nor did it always solve important problems of interest. Major questions were raised: What does it take to make GP a science? What does it take to make GP a technology? How to get as many people to use GP as possible; reducing up-front human effort; scalability as a moving target (scales just keep growing). 

These problems and others, along with success stories, showed at least one thing: There was definitely a future for the GPTP workshop series (and for GP in general). 

\paragraph{GPTP IX: 2011.}
\citet{Riolo:2011} observed once again important themes from previous years: modularity, scalability, evolvability, and the theoretical understanding of GP complexity and operators. Half of the chapters this year were devoted to symbolic regression, a mainstay of GPTP, which has persisted to this day as a major application triumph of the field of GP. 

Concern was expressed that while symbolic regression was perhaps a mature subfield, and the most successful application area of GP, it might cause the field to stagnate. There were still many real challenges in the original GP arena. There was even a suggestion to ban symbolic regression the following year, to make room for other areas (the ban did not go through \ldots). 

Participants of the 2011 workshop were asked to provide detailed feedback regarding the workshop series, and many interesting topics thus came up, a (very) partial list of which includes: Using GP to evolve software in general was cited as a potential future breakthrough; novelty search was mentioned by several participants as an interesting development \cite{Lehman2008}; a grand challenge posed was explaining why GP works for some problems and fails for others. There was also a concern raised that GP had trouble with certain classes of complex problems.

\citet{Riolo:2011} noted that there was consensus as to the lack of morphogenesis at the workshop. Moreover, there did not seem to be much progress in attaining open-ended evolution nor in understanding evolvability.

\paragraph{GPTP X: 2012.}
\citet{Riolo:2012} pointed out an important change in the tenth edition of the workshop: a more varied mix of different representations of GP individuals, e.g., SQL query, image filter, power control algorithm, and game-board evaluation function. Moreover, a conscious effort was made to go beyond GP symbolic regression, an application that starred in previous events.

GP's success as a field was unequivocally agreed upon: 
``Another positive observation is that the existential discussions on whether GP can declare success as a science have dissipated from GPTP. The overall consensus is that GP has found its niche as a capacious and flexible scientific discipline, attracting funding, students, and demonstrating measurable successes in business.'' A bumpy yet successful 10-year ride.

Focus shifted from comparing GP with other disciplines hoping to come out ahead, towards a more productive search for high-impact problems solvable with some form of GP. 

\citet{Riolo:2012} remarked on the increasing gap between theory and practice, expressing doubt it would ever be closed. In addition to necessary continued research into theory, several other directions were raised by the participants, including: what problems can be solved using a computer program and how can GP be used therein; how to seamlessly integrate different types of data structures; various requirements for open-ended evolution; exploiting modern architectures to run GP; post-processing and selection of GP solutions.

\paragraph{GPTP XI: 2013.} 
A regrettable development beginning with this workshop \cite{Riolo:2013} is the lack of an insightful, extended summary chapter, included in the first ten volumes, which contained priceless accounts of oral discussions. 
This places limits on our account henceforth, but we have still been able to glean valuable insight.

Symbolic regression once again emerged (unsurprisingly) as a main topic of interest. Designing GP systems to tackle real-world problems continued to be at the forefront, e.g., genetic analysis of Alzheimer's disease, designing Flash memory, and explaining unemployment rates. A novel genetic operator was presented (incorporating aspects of both crossover and mutation) and semantic GP (integrating semantic awareness in the genetic operators) was demonstrated for real-life applications. 

This edition of GPTP seemed to be mostly about pushing the application boundaries further.

\paragraph{GPTP XII: 2014.}
This workshop saw the rise of big data and machine learning, evident in the keynote talks \cite{Riolo:2014}. In addition, the format of the workshop was expanded to include ``whiteboard sessions''---discussions led by a participant on a topic of ongoing research. ``The goal of these sessions was to expand the speculative nature of the workshop so that people could present and discuss topics on which their thinking was not yet complete in order to solicit new ideas and suggestions on these topics.'' \cite{Riolo:2014preface}
Three such sessions took place: evolving arbitrary software, mobile computing and evolutionary computing, and application spaces and opportunities. 

Three of the ten chapters dealt with symbolic regression. Of note amongst the rest was the analysis of a decade of human-competitive (``HUMIE'') winners \cite{Kannappan:2015}, a trend whose seedlings we noted above in the very first editions of GPTP. GP was now producing impressive results on a routine basis.

\paragraph{GPTP XIII: 2015.}
The keynotes at this workshop  were prescient in that they admitted two topics that have since gained prominence: autonomous vehicles and games \cite{Riolo:2015}. Symbolic regression, comprising four of the fourteen chapters, clearly continued to advance. 

Massively parallel GP systems suitable for cloud computing were now of interest.
With the rise of data science, a special chapter was devoted to the use of GP in this field. Another interesting topic was a speculation on the possible use of open-ended GP in the Internet of Things; evolution was seen to be the major force behind the putative singularity. A novel selection scheme was discussed (lexicase) as well as the use of graph databases to explore the dynamics of GP runs. These latter complemented the advanced application talks, showing that the boundaries of fundamental GP (operators, dynamics) were also being explored.

\paragraph{GPTP XIV: 2016.}
This fourteenth edition saw more papers on advanced applications, with a larger presence of big data and machine learning \cite{Tozier:2016} . 
Differences between biological and computational evolution came up yet again in one of the keynote talks. Another keynote focused on the newly emerging, important field of genetic improvement and automated software repair.
A prediction made by \citet{Orlov:2010} back in GPTP VIII seemed to have progressed towards reality: ``We believe that in about fifty years' time it will be possible to program computers by means of evolution. Not merely \textit{possible} but indeed \textit{prevalent}.''

An interesting talk presented the use of GP to attain artificial general intelligence in video games. 
The HeuristicLab system showcased a user-friendly, heuristic and evolutionary algorithm framework.
There were at least three chapters that 
arose from many years of fruitful exploration of the Push language for GP (a programming language designed for use in auto-constructive evolution of computer programs).
Also tackled were some major issues mentioned above: diversity, robustness, and evolvability. 

\paragraph{GPTP XV: 2017.}
This event boasted three keynote talks, covering a very up-to-date spectrum of topics: the forefront of modern AI,
open-ended evolution, and the collaboration of humans and computers to extract information and produce knowledge from data \cite{Banzhaf:2017}. In line with GPTP's goal of extending the boundaries of fundamental GP, lexicase selection again made an appearance as a powerful alternative to classical selection methods.

Another interesting presentation focused on 
evolving novel solutions to real-world problems through nature-like ecological dynamics that generate a diverse array of raw materials for evolution to build upon. 
An industrial example of financial investments based on GP was hailed as a categorical success. A GP schema-analysis paper harked back to the very origins of this workshop series. 
Machine learning and artificial intelligence were present in force, with one ambitious paper presenting a system for accessible, open-source, user-friendly AI.

While this essentially concludes our 15-year trajectory we would like to briefly mention GPTP XVI, which took place in May 2018 (the proceedings are still in preparation).
Machine learning continued its rise to prominence, with a full one third of the talks being in this area.
Two papers focused on symbolic regression, and others involved well-known challenges, encountered above: operators (e.g., selection), developmental GP, and coevolution. 

\section{Discussion and Concluding Remarks}

GPTP began with a wide divide between theory and practice. The first few workshops dealt with ``classical'' issues in theory---operators, building blocks etc.---along with applications of GP. By the third workshop both groups---theoreticians and practitioners---were showing signs of convergence, each influencing the other. The early workshops saw the beginning of human-competitive results produced by GP.

By the fourth workshop two points were clear: 1) GP could solve real-world problems but 2) not by using ``pure'' GP. Since then, much of GPTP has focused on advanced, best-practice techniques needed for GP to solve complex problems---both from a theoretical standpoint of understanding why certain approaches worked, and from a practical perspective of making headway into complex territory.

The use of GP by non-experts was a recurring theme in many of the earlier events, but at some point these discussions subsided. Apparently, researchers realized that time was best spent on solving complex problems using GP expertise, rather than attempting to build a basic GP system that could be used by a novice to solve simple problems. There emerged a ``kitchen sink'' approach, wherein advanced techniques were added both by theoreticians and by practitioners to gain ground in solving hard problems.

By GPTP VII there was a consensus that GP was routinely (``a watershed'') delivering solutions to highly complex problems, with best-practice approaches proving very successful. The following year a new trend started to emerge, that of GP-enhanced, user-friendly, usable \textit{systems}. As in the previous events, major topics repeatedly encountered included efficient use of computational resources, ensuring quality results, and reliable convergence. By GPTP VIII toy problems had disappeared altogether in favor of real-world problems and GP systems.

Symbolic regression persists to this day as arguably the biggest success of GP in real-life applications. Nary a year went by without at least one paper on this topic  (usually more). So much so that at some point concern was raised that focusing on symbolic regression would lead to stagnation in GP research. Indeed, by GPTP X there seemed to be a conscious effort to move beyond symbolic regression, with a plethora of novel representations cropping up. Moreover, by this tenth edition a consensus was apparent regarding GP's unequivocal success at solving hard, real-world problems. Also, research was starting to focus on success in high-impact problems rather than on simply comparing with other approaches.

GPTP originally had the stated intent of closing the gap between theoreticians and practitioners. Ten years onward it was evident the gap would persist but continued research on the frontier would help both camps. 

By GPTP XII human-competitive results were being routinely produced by GP, enough to warrant a ten-year overview. The following year saw GP researchers starting to foray into big data and machine learning, a trend that has continued (unsurprisingly). The very latest events have seen more machine learning and data science topics, as well as a continual effort to produce both fundamental, novel insights (e.g., new operators) and better best practices for hard problems.

We have enumerated along the way many themes and challenges raised by the participants over the years. Below we summarize major recurring ones:
\begin{itemize}
\item GP is superb at symbolic regression---and this application is at the heart of many complex, real-world problems.
\item In real life one must progress beyond pure, vanilla GP.
\item Best practices (design of fitness, operators, dynamics, etc.) for hard problems.
\item Expert knowledge is needed. We do not yet possess a turnkey GP system that can tackle real-world problems.
\item Modularity, scalability, evolvability, and the theoretical understanding of GP complexity and operators.
\item Open-endedness and novelty search.
\item Development as an important phase in nature that should inspire work into genotype-to-phenotype mappings in GP.
\item Extended use of cluster and cloud computing.
\item Usability and user friendliness. 
\item Automatic programming and software evolution. 
\end{itemize}







\bibliographystyle{spmpscinat}
\bibliography{gptp-trajectory}   

\end{document}